\documentclass[runningheads]{llncs}

\usepackage{pgf,tikz}
\usepackage{verbatim}
\usepackage{graphicx}
\usepackage{siunitx}
\usepackage[T1]{fontenc}
\usepackage[utf8]{inputenc}
\usepackage[autostyle]{csquotes}

\newcommand{\ignore}[1]{}

\begin{document}

%\title{Toxicity prediction using multi-data and multi-model guided deep learning \thanks{Supported by %Institute x}}
%\title{Toxicity prediction using multi-data and multi-model guided deep learning}
\title{Toxicity Prediction by Multimodal Deep Learning}

%\begin{comment}
\author{Abdul Karim\inst{1} \and
Jaspreet Singh\inst{1} \and
Avinash Mishra\inst{2} \and
Abdollah Dehzangi\inst{3} \and
M. A. Hakim Newton\inst{4} \and
Abdul Sattar\inst{4}}
\authorrunning{Abdul Karim et al.}
\institute{School of Information Communication Technology, Griffith University, Australia
\email{Abdul.karim@griffithuni.edu.au, jaspreetsingh2@griffithuni.edu.au}\\
\and
Department of Chemical Engineering
Indian Institute of Technology
Hauz Khas, New Delhi 110016
India
\email{avish2k@gmail.com}\\
\and
Department of Computer Science, Morgan State University, Baltimore, USA\\
\email{abdollah.dehzangi@moegan.edu}
\and
Institute of Integrated and Intelligent Systems, Griffith University, Australia\\
\email{mahakim.newton@griffith.edu.au, a.sattar@griffith.edu.au}
}
\maketitle

\vspace{-2em}
\begin{abstract} 
Prediction of toxicity levels of chemical compounds is an important issue in Quantitative Structure-Activity Relationship (QSAR) modeling. Although toxicity prediction has achieved significant progress in recent times through deep learning, prediction accuracy levels obtained by even very recent methods are not yet very high. We propose a multimodal deep learning method using multiple heterogeneous neural network types and data representations. We represent chemical compounds by strings, images, and numerical features. We  train fully connected, convolutional, and recurrent neural networks and their ensembles. Each data representation or neural network type has its own strengths and weaknesses. Our motivation is to obtain a collective performance that could go beyond individual performance of each data representation or each neural network type. On a standard toxicity benchmark, our proposed method obtains significantly better accuracy levels than that by the state-of-the-art toxicity prediction methods.   
\vspace{-1ex}
\end{abstract}

\vspace{-2em}
\section{Introduction}
\vspace{-1ex}
Every year a broad spectrum of chemical compounds are produced in various laboratories all over the world. A large number of these chemical compounds are suspected to be toxic or hazardous for human life, and at the end, many of them are proven so. As a result, {\em toxicity prediction} has become one of the most important issues in Quantitative Structure-Activity Relationship (QSAR) modeling \cite{karim2019efficient,wu2018moleculenet}. Various functional groups and their specific three dimensional orientations make chemical compounds toxic in nature. The principal metric used for the measurement of toxicity is the concentration of compounds and the time of exposure to humans \cite{mcfarland1970parabolic}. The concentration of compounds that cause toxic or hazardous effect on human health are measured by experiments and are considered as {\em endpoints}. The exposure of toxic compounds to humans can take place through oral or intravenous uptake or inhalation. There exist several toxicity metrics but the most popular one is IGC50 \cite{zhu2008combinatorial}. IGC50 measures the concentration of the compounds that inhibit 50\% of growth on test population.

QSAR modelling has made significant progress in recent years through deep learning \cite{kato2016molecular}. To predict molecular activities via computational models, molecules are usually represented as strings of a given textual language such as Simplified Molecular-Input Line-Entry System (SMILES) \cite{bjerrum2017smiles}. Such SMILES strings can then be used to compute various types of numerical features (e.g. physicochemical descriptors) and molecular images \cite{yap2011padel}. Numerical features have been used in various traditional machine learning approaches such as K-Nearest Neighbours (KNN), Support Vector Machines (SVM), Random Forest (RF), and Fully Connected Neural Networks (FCNN) \cite{lima2016use}. On the other hand, molecular images have been used in Convolutional Neural Networks (CNN) \cite{goh2018much}. Computation of molecular images needs relatively less domain specific expertise than that of numerical features, but CNN models using them still achieve reasonable performance levels \cite{goh2018much} compared to the other models using numerical features. SMILES strings can also be transformed into a vector representation and used in Recurrent Neural Networks (RNN) for molecular activity prediction \cite{goh2018smiles2vec}.     

In recent work on toxicity prediction, physicochemical descriptors and fingerprints are used in deep neural networks and consensus models by TopTox \cite{wu2018quantitative} to predict regression activity such as Pearson correlation coefficient $R^2$ between the experimental and predicted toxicity levels. Another system named AdmetSAR \cite{yang2018admetsar} uses molecular fingerprints to predict $R^2$ values by RF, SVM, and KNN models. Yet another system referred to here by the name Hybrid2D \cite{karim2019efficient} uses a hybridization of shallow neural networks and decision trees on 2D features only to predict $R^2$ values. TopTox, AdmetSAR, and Hybrid2D use an IGC50-based benchmark dataset as one of their benchmarks and obtain accuracy levels 0.80--0.83 on that dataset. Clearly, these are not very high accuracy levels.

In this paper, we propose a {\em multimodal deep learning method} that uses multiple {\em heterogeneous} neural network types and data representations. We represent the formula of a chemical compound as a SMILES string and as a molecular image. We further represent the chemical compound using numerical features obtained from physicochemical descriptors. We train an RNN on vector representations of SMILES strings, FCNN on numerical feature values, and CNN on molecular images. We then build an ensemble from the RNN, the FCNN, and the CNN using an Ensemble Averaging (EA) method or a Meta Neural Network (MNN) to obtain the final output. Each data representation type or each neural network type has its own strengths and weaknesses. Our motivation is to obtain a collective performance that could go beyond the individual performance of each data representation or each neural network type. Our multimodal approach is different from a typical ensembling approach as the latter uses homogeneous neural networks and data representations. On the IGC50 toxicity benchmark dataset, our proposed method obtains significantly better accuracy levels (0.84--0.88) than that by the state-of-the-art toxicity prediction methods TopTox, AdmetSAR, and Hybrid2D.  
  
In the rest of the paper, Section~\ref{secPreliminaries} covers preliminaries of toxicity prediction and neural networks, Section~\ref{secMethodologies} describes our multimodal deep learning approach, Section~\ref{secResults} provides experimental results, and Section~\ref{secConclusions} presents conclusions. 

\section{Preliminaries\label{secPreliminaries}}

We give overviews of SMILES strings, the IGC50 dataset, and neural networks.

\subsection{SMILES Strings}

SMILES is a text-based chemical language that is used to describe the information about the structure of a molecule in a single line of characters \cite{weininger1988smiles}. SMILES strings obey a regular grammar or syntax. Various types of characters are used to denote atoms and bonds between them. For example, c is used for representing aromatic carbon whereas C represents aliphatic carbon. There are special characters like \enquote{=} and \enquote{-} to denote double and single bonds respectively. An example of a SMILE string is \enquote{CC1=CC(=O)C2=C(C=CC=C2O)C1=O}.

\subsection{IGC50 Dataset}

Among several toxicity metrices, IGC50 is one of the most important endpoints \cite{zhu2008combinatorial}. IGC50 measures the concentration of compounds that inhibit 50\% of growth on test population. The benchmark dataset, denoted henceforth by IGC50 dataset and used in this work, has IGC50 values and their test population is Tetrahymena Pyriformis \cite{wu2018quantitative}. Tetrahymena Pyriformis is an aquatic animal (Protozoa) that lives in fresh water. It is pear-shaped, $50\times 30$ pm in length, multiplies in 3h to 4h and can be cultured in a single membered sterile culture \cite{new4,new5}. Thus, IGC50 in the given dataset refers to acute aquatic toxicity of compound on Tetrahymena Pyriformis population. The time of exposure considered here is 40h, which indicates that population of Tetrahymena Pyriformis are exposed to these compounds for 40h and then reduction in growth was measured \cite{wu2018quantitative}. IGC50 values reported in the given dataset is measured in $-\log_{10}(C)$ where $C$ is the concentration in $\textrm{mol}/\textrm{L}$ \cite{wu2018quantitative}. There are 1792 compounds in the IGC50 dataset. These compounds are represented as SMILES strings with lengths ranging from 2 to 52 characters.

\subsection{Neural Networks}
A {\em deep neural network} (DNN) has multiple hidden layers while a {\em shallow neural network} (SNN) typically has only one hidden layer. We refer the reader to \cite{schmidhuber2015deep} for the concepts and mathematics of deep learning on DNNs. Below we briefly cover various types of neural networks based on their architectures.

\begin{enumerate}
\item {\bf FCNN.} A neural network in which each unit of one layer is connected to all units of the next layer is termed as a {\em fully connected neural network} (FCNN). FCNNs take numerical features as an input to predict the output. 
\item {\bf CNN.} A {\em convolutional neural network} is a special type of neural network for the image data. CNNs can extract low level features from images and compute more complex features as we go deeper in the networks \cite{szegedy2015going}. Variants of CNN like Inception, Alexnet and Resnet have been developed and employed as highly accurate image classification models \cite{he2016deep}. 
\item {\bf RNN.} A {\em recurrent neural network} is a specialized neural network for sequential data. RNNs can learn features directly from the sequence data without explicitly computing features. RNNs use their internal state (memory) to process the sequence of data. They have shown great success in natural language processing and machine translation \cite{mikolov2013efficient}. RNNs usually are prone to short term memory problem \cite{hochreiter1997long}. The information flows from one cell to another sequentially and might be corrupted later in the network for longer sequences. Long short-term memory (LSTM) units or gated recurrent units (GRU) in RNN offer solutions to the short term memory problem \cite{cho2014learning}.
\item {\bf Ensembles.} An {\em ensemble} is a collection of multiple {\em component neural networks}. {\em Ensemble averaging} (EA) is a method to average out the outputs of multiple component neural networks in an ensemble. A {\em meta neural network} (MNN) may also be used for averaging out. Ensembles of neural networks often perform better than individual neural networks. Usually the data representations and the network types (e.g. FCNN or CNN or RNN) of all the neural networks in an ensemble are the same. An MNN if used is normally a shallow FCNN. We assume the FCNN, CNN, or RNN component neural networks used in ensembles are deep neural networks.
\end{enumerate}

\section{Methodologies\label{secMethodologies}}

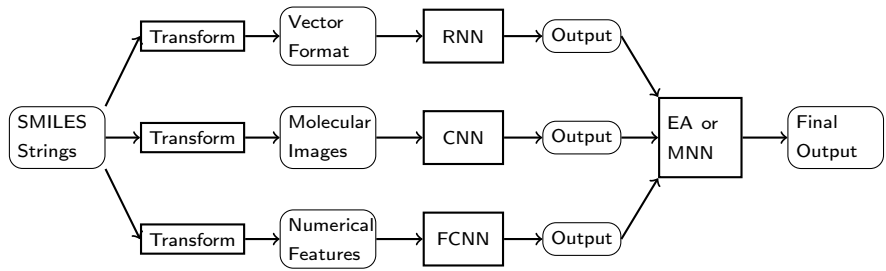
\begin{figure}[!hptb]
\centering
\begin{tikzpicture}[scale=0.9]
\node (ss) at (-5,0) [draw,text width=30,rounded corners=5] {\sf\scriptsize SMILES Strings};

\node (fcin) at (-3,-1.5) [draw,thick] {\sf\scriptsize Transform};
\node (cin) at (-3,0) [draw,thick] {\sf\scriptsize Transform};
\node (rin) at (-3,1.5) [draw,thick] {\sf\scriptsize Transform};

\draw[->,thick] (ss.south east) -- (fcin.west);
\draw[->,thick] (ss.east) -- (cin.west);
\draw[->,thick] (ss.north east) -- (rin.west);

\node (tdd) at (-1,-1.5) [draw,text width=30,rounded corners=5] {\sf\scriptsize Numerical Features};
\node (mi) at (-1,0) [draw,text width=30,rounded corners=5] {\sf\scriptsize Molecular Images};
\node (vf) at (-1,1.5) [draw,text width=30,rounded corners=5] {\sf\scriptsize Vector Format};

\draw[->,thick] (fcin.east) -- (tdd.west);
\draw[->,thick] (cin.east) -- (mi.west);
\draw[->,thick] (rin.east) -- (vf.west);

\node (fcnn) at (1,-1.5) [draw,thick,minimum width=30,minimum height=20] {\sf\scriptsize FCNN};
\node (cnn) at (1,0) [draw,thick,minimum width=30,minimum height=20] {\sf\scriptsize CNN};
\node (rnn) at (1,1.5) [draw,thick,minimum width=30,minimum height=20] {\sf\scriptsize RNN};

\draw[->,thick] (tdd.east) -- (fcnn.west);
\draw[->,thick] (mi.east) -- (cnn.west);
\draw[->,thick] (vf.east) -- (rnn.west);

\node (fcout) at (2.75,-1.5) [draw,rounded corners=5] {\sf\scriptsize Output};
\node (cout) at (2.75,0) [draw, rounded corners=5] {\sf\scriptsize Output};
\node (rout) at (2.75,1.5) [draw, rounded corners=5] {\sf\scriptsize Output};

\draw[->,thick] (fcnn.east) -- (fcout.west);
\draw[->,thick] (cnn.east) -- (cout.west);
\draw[->,thick] (rnn.east) -- (rout.west);

\node (ea) at (4.5,0) [draw,thick,minimum height=30,text width=25] {\sf\scriptsize EA or MNN};
\draw[->,thick] (fcout.east) -- (ea.south west);
\draw[->,thick] (cout.east) -- (ea.west);
\draw[->,thick] (rout.east) -- (ea.north west);

\node (fout) at (6.5,0) [draw,text width=30,rounded corners=5] {\sf\scriptsize Final Output};

\draw[->,thick] (ea.east) -- (fout.west);
\end{tikzpicture}
\caption{Our proposed multimodal deep learning architecture for toxicity prediction\label{figMultimodal}}
\end{figure}

Our multimodal deep learning method uses multiple heterogeneous neural network types and data representations within an ensemble of neural networks. \figurename~\ref{figMultimodal} shows the proposed multimodal deep learning architecture. SMILES strings of chemical compounds are first transformed into a vector format, or a molecular image format, or a set of numerical features. Then, an RNN, a CNN, and an FCNN are trained respectively on the vector format, image format, and the numerical features. The coupling between the data representations and the neural network types are because the respective neural networks are the best suited ones for the respective data representations. The outputs of the component RNN, CNN, and FCNN are the averaged out through an EA method or using an MNN to obtain the final output. We further describe each part of the architecture.     

\subsection{Vector Representation}

Each character of a SMILES string is represented by a 50 component one-hot vector, where only one bit is high and all other bits are low.

\vspace{-1em}

\subsection{Molecular Images}

\begin{figure}[h!]
 \vspace{-1em}
 \centering
	 \includegraphics[width=\textwidth]{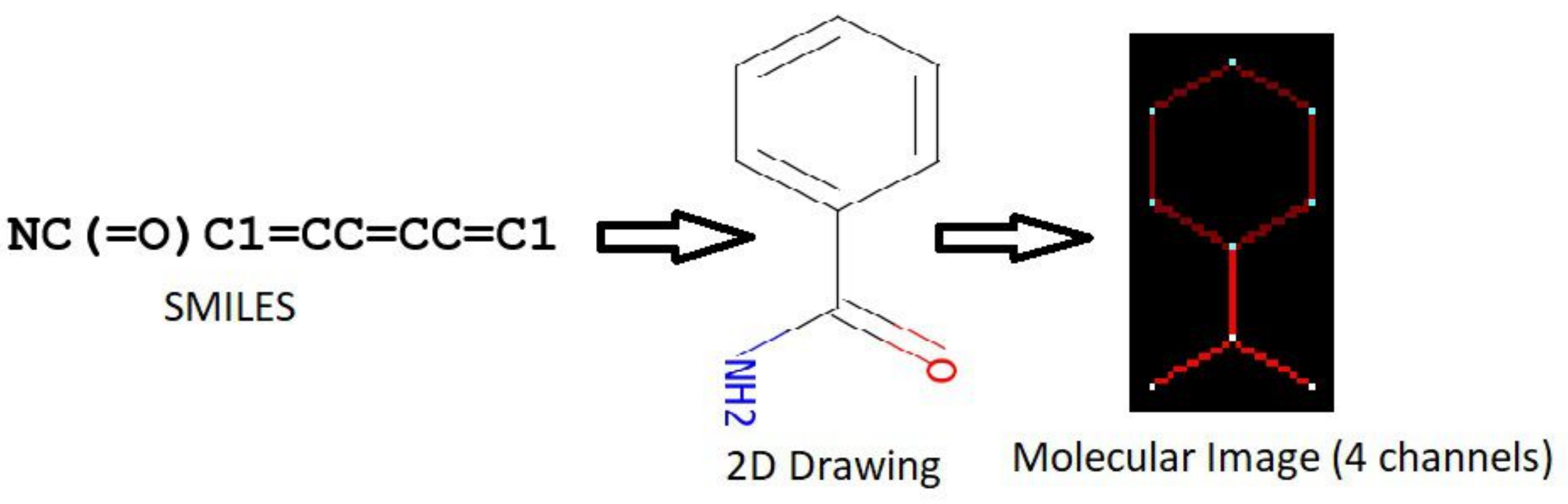} 
  \caption{Computing a molecular image from 2D coordinates generated from a SMILES string by using an open source python library rdkit}
  \label{fig:Images}
\end{figure}

SMILES strings are used to generate 2D molecular images \cite{goh2018much}; see \figurename~\ref{fig:Images}. An open source python library rdkit is used to generate 2D drawings of the SMILES strings in the IGC50 dataset \cite{landrum2013rdkit}. The 2D coordinates are mapped onto a grid of size $100\times 100$ with a pixel resolution of $0.2\si{\angstrom}$. Depending upon the presence of bonds or atoms, the gray scale images are color coded with 4 channels. Each channel encode different information about the molecule. Layer zero is used for the information about the bonds and the other three layers are for atomic numbers, gasteiger charges, and hybridization.

\vspace{-2em}
\subsection{Numerical Features}

2D numerical features used are less multifarious in nature and easy to calculate. 1422 2D features are computed using an open source software PADEL descriptor \cite{yap2011padel}. The main reason for using 2D features is that these descriptors have shown promising prediction power in a previous study \cite{karim2019efficient}. 

\subsection{Input Output}

All the three types of input data generated from the SMILES strings in the IGC50 dataset are fed into three types of suitable deep learning approaches to predict Pearson correlation coefficient $R^2$ values. 

\subsection{FCNN}

We use a neural network with two hidden layers, each consisting of 100 units. The training data size is 1792 molecules with 1422 2D numerical features as described before. A random optimization technique {\bf REF} is used to obtain the optimized values of the neural network parameters as shown in Table~\ref{tab1}. Adam optimization with default learning rate is used as the back propagation gradient descent \cite{kingma2014adam}. The drop out is used after first hidden layer only.
\begin{table}[!h]
\centering
\caption{Optimized parameters for FCNN}
\label{tab1}
\begin{tabular}{|l|l||l|l|}
\hline
Parameter Name & Parameter Value & Parameter Name & Parameter Value\\
\hline
Epochs &  400  & Initialization Function &  Glorot-Normal \\
DropOut & 0.1  & Activation (1st layer) & Sigmoid \\
Mini-batch & 1024 & Activation (2nd layer) & Relu\\
\hline
\end{tabular}
\vspace{-2em}
\end{table}

\subsection{CNN}
\begin{figure}[h!]
	\vspace{-2em}
  \centering
  \includegraphics[width=\linewidth]{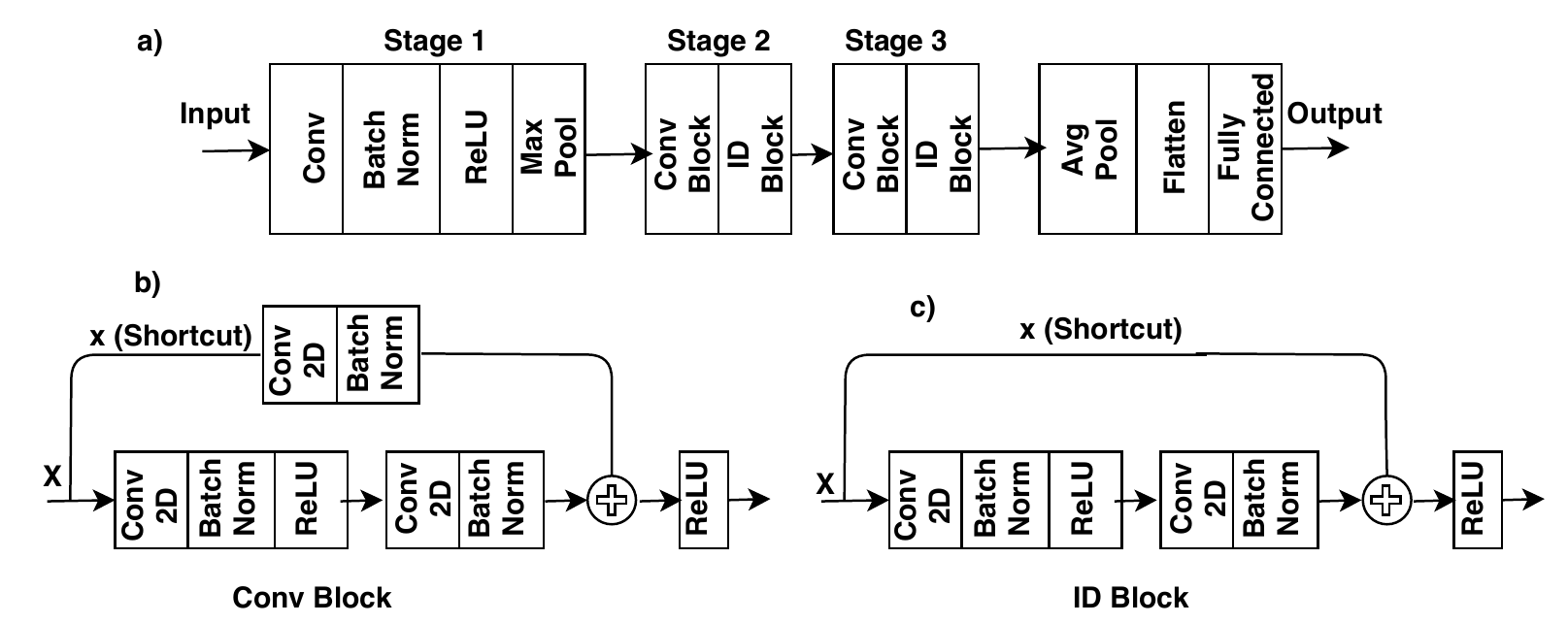} 
  \caption{Resnet architecture used in CNN}
  \label{fig:Resnet}
\end{figure}

We use a three stage Resnet as shown in \figurename~\ref{fig:Resnet}a. The Resnet consists of residual connections (skip connections), which make it prone to the vanishing gradient problem \cite{he2016deep}. It allows the gradient to propagate to the early layer without vanishing. This type of skip connection is inherited in convolutional block and identity blocks in the network as shown in \figurename~\ref{fig:Resnet}b and c. Adam optimizer with default learning rate and 128 batch size are used. The number of epochs is 150 with an early stopping criterion. The implementation detail of each layer is given below. 

\vspace{-1em}

\begin{itemize}
\item[$\bullet$] {\bf Input:} Input image is of the shape ($100\times 100$) with 4 channels.
\item[$\bullet$] {\bf Stage 1:} The 2D convolution has 64 filters of shape (7, 7) and uses a stride of (2, 2). BatchNorm is applied to the channels axis of the input. MaxPooling uses a (3, 3) window and a (2, 2) stride.
\item[$\bullet$] {\bf Stage 2:} The convolutional block uses three set of filters of size [64, 256, 256] each with a shape (1, 1) and stride (1, 1). The identity block use two sets of filters of size [64,256] each with a shape (1, 1) and stride (1, 1).
\item[$\bullet$] {\bf Stage 3:} The convolutional block uses three set of filters of size [128, 512, 512] each with a shape (1, 1) and stride (1, 1). The identity block use two sets of filters of size [128, 512] each with a shape (1, 1) and stride (1, 1)
\item[$\bullet$] {\bf Average pooling:} The 2D average pooling uses a window of shape (2, 2).
\item[$\bullet$] {\bf Flatten:} It is a function that converts the pooled features from the max pooling layer into a single column feature vector. 
\item[$\bullet$] {\bf Fully connected:} A dense layer which is fully connected to the previous single column vector generated by flatten. For a regression problem like in case of IGC50 molecular images, it consists of single neuron or unit. 
\end{itemize}

\subsection{RNN}

\begin{figure}[!hptb]
	\vspace{-1em}
	\centering
	\begin{tikzpicture}[scale=0.9]
		\node (node1) at (-0.5,0) [minimum height=30,text width=30] {\sf\scriptsize One-Hot Vectors};
		\node (node2) at (2,0) [draw,minimum height=30,text width=40] {\sf\scriptsize Embedding Layer};
		\node (node3) at (5,0) [draw,minimum height=30,text width=40] {\sf\scriptsize 3 $\times$ 1D Convolution};
		\node (node4) at (7.5,0) [draw,minimum height=30,text width=40] {\sf\scriptsize Flatten};
		\node (node5) at (10,0) [draw,minimum height=30,text width=40] {\sf\scriptsize Fully Connected};
		\draw[thick,->] (node1) -- (node2);
		\draw[thick,->] (node2) -- (node3);
		\draw[thick,->] (node3) -- (node4);
		\draw[thick,->] (node4) -- (node5);
	\end{tikzpicture}
  \caption{RNN architecture}
  \label{fig:RNN}
\end{figure}
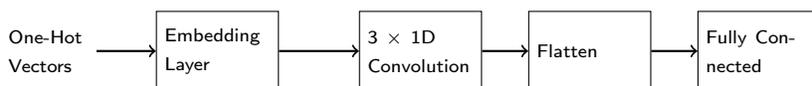

 We developed a variant of RNN which involves 1D convolutions instead of LSTM or GRU as shown in Figure \ref{fig:RNN}. The reason of using 1D convolution instead of GRU or LSTM is because IGC50 molecules are shorter in length. All the unique SMILES characters in the sequence are mapped to integer numbers using a dictionary. One-hot vector encoded characters are fed into a network. An embedding layer is used to compute an embedded vector representation of SMILES sequence. It should be noted that ReLu activation function is used with convolution layers while linear activation function is used with fully connected or dense layer. Adam optimizer with default learning rate and 128 batch size is used. The number of epochs is 150 with an early stopping criterion. The implementation detail of the RNN architecture in Figure \ref{fig:RNN} is given below.

\vspace{-1em}
\begin{itemize}  
\item[$\bullet$] {\bf One-hot vectors:} Every character of each SMILES string is one hot vector encoded and fed into embedded layer. 
\item[$\bullet$] {\bf Embedding layer:} One-hot vectors for 50 dimensional space.
\item[$\bullet$] {\bf 1D convolution layer:} Each 1D convolution is performed using 92 filters with size of 10, 5 and 3 respectively.
\item[$\bullet$] {\bf Flatten:} A function that flatten out the output of 1D convolution.
\item[$\bullet$] {\bf Fully connected or dense:} The fully connected layer computes the output. It is densely connected all neurons from the previous layer.
\end{itemize}

\subsection{EA or MNN}

Each of the component FCNN, CNN, and RNN is trained independently. When the EA method is used, the final output is the average of the output of the component neural networks. When an MNN is used, we consider the outputs of the FCNN, CNN, and RNN as three input features to the MNN and then train the MNN. The MNN has only one hidden layer with 10 neurons. We use Adam optimizer with the default learning rate to optimise the MNN. Also, we use 400 epochs and an early stopping criterion. After performing hyper-parameter random search, we use  mini-batch size of 512, drop-out of 0.4, glorot-normal initialization function and sigmoid activation.

\subsection{Implementation}
All the neural network models are built using a Keras deep learning framework on a system with NVidia Tesla K40 GPU.

\section{Results\label{secResults}}

We split the data into train(70\%) and test(30\%) sets randomly in the beginning of modeling. The test set is kept aside (blind) for the final testing after finalizing the hyper-parameters like epoch, drop-out, activation function, mini-batch size and initialization function using 5 fold cross-validation (CV) on the train set. Table~\ref{tabResults} presents the $R^2$ values obtained by component neural works, their ensembles, and the existing state-of-the-art methods.

\begin{table}[!hptb]
\vspace{-1em}
\centering
\caption{Performance comparison on ($R^2$) values using IGC50 dataset\label{tabResults}}
\label{tab3}
\begin{tabular}{|l||l|l|l||l|l||l|l|l|}
\hline
		& FCNN & CNN & RNN & EA & MNN & TopTox & AdmetSAR & Hybrid2D\\\hline
CV		& \underline{0.82}	& 0.80 & 0.78 & 0.85 & \bf 0.88 & NA & 0.82 & 0.83\\
Test 	& \underline{0.81} 	& 0.78 & 0.79 & 0.84 & \bf 0.86 & 0.80 & NA & 0.81\\\hline
\end{tabular}
\vspace{-2em}
\end{table}

\subsection{Component Neural Networks}

FCNN achieves better performance than CNN and RNN on test and CV. For CV, FCNN achieves 2\% better accuracy than CNN and 4\% better than RNN. For test, FCNN outperforms CNN and RNN base model by 3\% and 2\% respectively.

\subsection{Ensemble Performance}

For CV, the EA method improves the ($R^2$) value to 0.85 whereas the MNN approach improves it to 0.88. For test, the EA method improves the ($R^2$) value to 0.84 whereas the MNN approach improves it to 0.86.

\subsection{Existing Methods}

We compare the performance of our proposed methods with three state-of-the-art toxicity prediction methods. These three methods are described below.

\begin{enumerate}
\item {\bf TopTox} \cite{wu2018quantitative} uses various types of approaches such as single task deep neural network, multi-task deep neural network and consensus models to verify the predictive power of element specific topological descriptors, auxiliary molecular descriptors (AUX), and a combination of both.

\item {\bf AdmetSAR} \cite{yang2018admetsar} represents molecules by fingerprints such as MACCS, Morgan and AtomParis implemented with RDKit. Machine learning algorithms including RF, SVM, and KNN are used to build the models.

\item {\bf Hybrid2D} \cite{karim2019efficient} is using hybrid optimization of shallow neural network and decision trees to prerdict $R^2$ values using only 2D Features.
\end{enumerate}

As we see from Table~\ref{tabResults}, performance of our ensembled approaches are better than that of all the three existing methods both on CV and test.

\subsection{Analyses and Discussions}

From the results in Table~\ref{tabResults}, it appears interesting that RNN with the vector representation of just SMILES strings and CNN with molecular images obtain similar performances on IGC50 datasets. It raises the question as to the usefulness of the CNN with molecular images. We leave this for future study. While ensembles improve performance over component neural networks, the MNN approach appears to be better than the EA approach.

We selected the IGC50 dataset which has relatively small compounds compared to the other datasets. This is because large molecules are difficult to encode in fixed sized 2D molecular images. We leave it for future study to use some other datasets or using some other data representations.

\section{Conclusions\label{secConclusions}}

Multimodal data representations and network types best suited to the data representations can capture various aspects of a machine learning task. In this paper, we propose a multimodal deep learning method that uses multiple heterogeneous neural network types and data representations. We represent the formula of a chemical compound in a textual language, in an image format and also in terms of numerical features. We then build an ensemble from various types of deep neural networks suitable for the data representations. Our multimodal approach is different from a typical ensembling approach as the latter uses homogeneous neural networks and data representations. On the IGC50 toxicity benchmark dataset, our proposed method obtains significantly better accuracy levels (0.84--0.88) than that (0.80--0.83) by the state-of-the-art toxicity prediction methods.  

\section*{Acknowledgment}

We gratefully acknowledge the support of NVIDIA Corporation with the donation of the Titan XP GPU used for this research.
\bibliography{pkaw2019}

\begin{thebibliography}{10}
\providecommand{\url}[1]{\texttt{#1}}
\providecommand{\urlprefix}{URL }
\providecommand{\doi}[1]{https://doi.org/#1}

\bibitem{bjerrum2017smiles}
Bjerrum, E.J.: Smiles enumeration as data augmentation for neural network
  modeling of molecules. arXiv preprint arXiv:1703.07076  (2017)

\bibitem{cho2014learning}
Cho, K., Van~Merri{\"e}nboer, B., Gulcehre, C., Bahdanau, D., Bougares, F.,
  Schwenk, H., Bengio, Y.: Learning phrase representations using rnn
  encoder-decoder for statistical machine translation. arXiv preprint
  arXiv:1406.1078  (2014)

\bibitem{dietterich2002ensemble}
Dietterich, T.G., et~al.: Ensemble learning. The handbook of brain theory and
  neural networks  \textbf{2},  110--125 (2002)

\bibitem{new5}
Frankel, J.: Cell biology of tetrahymena thermophila. In: Methods in cell
  biology, vol.~62, pp. 27--125. Elsevier (1999)

\bibitem{goh2018smiles2vec}
Goh, G.B., Hodas, N., Siegel, C., Vishnu, A.: Smiles2vec: Predicting chemical
  properties from text representations. In: Workshop track, International
  Conference on Learning Representations (2018)

\bibitem{goh2018much}
Goh, G.B., Siegel, C., Vishnu, A., Hodas, N., Baker, N.: How much chemistry
  does a deep neural network need to know to make accurate predictions? In:
  2018 IEEE Winter Conference on Applications of Computer Vision (WACV). pp.
  1340--1349. IEEE (2018)

\bibitem{he2016deep}
He, K., Zhang, X., Ren, S., Sun, J.: Deep residual learning for image
  recognition. In: Proceedings of the IEEE conference on computer vision and
  pattern recognition. pp. 770--778 (2016)

\bibitem{new4}
Hill, D.G.: The biochemistry and physiology of Tetrahymena. Elsevier (2012)

\bibitem{hochreiter1997long}
Hochreiter, S., Schmidhuber, J.: Long short-term memory. Neural computation
  \textbf{9}(8),  1735--1780 (1997)

\bibitem{karim2019efficient}
Karim, A., Mishra, A., Newton, M.H., Sattar, A.: Efficient toxicity prediction
  via simple features using shallow neural networks and decision trees. ACS
  Omega  \textbf{4}(1),  1874--1888 (2019)

\bibitem{kato2016molecular}
Kato, Y., Hamada, S., Goto, H.: Molecular activity prediction using deep
  learning software library. In: 2016 International Conference On Advanced
  Informatics: Concepts, Theory And Application (ICAICTA). pp.~1--6. IEEE
  (2016)

\bibitem{kingma2014adam}
Kingma, D.P., Ba, J.: Adam: A method for stochastic optimization. arXiv
  preprint arXiv:1412.6980  (2014)

\bibitem{landrum2013rdkit}
Landrum, G.: Rdkit documentation. Release  \textbf{1},  1--79 (2013)

\bibitem{lima2016use}
Lima, A.N., Philot, E.A., Trossini, G.H.G., Scott, L.P.B., Maltarollo, V.G.,
  Honorio, K.M.: Use of machine learning approaches for novel drug discovery.
  Expert opinion on drug discovery  \textbf{11}(3),  225--239 (2016)

\bibitem{mcfarland1970parabolic}
McFarland, J.W.: Parabolic relation between drug potency and hydrophobicity.
  Journal of medicinal chemistry  \textbf{13}(6),  1192--1196 (1970)

\bibitem{mikolov2013efficient}
Mikolov, T., Chen, K., Corrado, G., Dean, J.: Efficient estimation of word
  representations in vector space. arXiv preprint arXiv:1301.3781  (2013)

\bibitem{schmidhuber2015deep}
Schmidhuber, J.: Deep learning in neural networks: An overview. Neural networks
   \textbf{61},  85--117 (2015)

\bibitem{szegedy2015going}
Szegedy, C., Liu, W., Jia, Y., Sermanet, P., Reed, S., Anguelov, D., Erhan, D.,
  Vanhoucke, V., Rabinovich, A.: Going deeper with convolutions. In:
  Proceedings of the IEEE conference on computer vision and pattern
  recognition. pp.~1--9 (2015)

\bibitem{weininger1988smiles}
Weininger, D.: Smiles, a chemical language and information system. 1.
  introduction to methodology and encoding rules. Journal of chemical
  information and computer sciences  \textbf{28}(1),  31--36 (1988)

\bibitem{wu2018quantitative}
Wu, K., Wei, G.W.: Quantitative toxicity prediction using topology based
  multitask deep neural networks. Journal of chemical information and modeling
  \textbf{58}(2),  520--531 (2018)

\bibitem{wu2018moleculenet}
Wu, Z., Ramsundar, B., Feinberg, E.N., Gomes, J., Geniesse, C., Pappu, A.S.,
  Leswing, K., Pande, V.: Moleculenet: a benchmark for molecular machine
  learning. Chemical science  \textbf{9}(2),  513--530 (2018)

\bibitem{yang2018admetsar}
Yang, H., Lou, C., Sun, L., Li, J., Cai, Y., Wang, Z., Li, W., Liu, G., Tang,
  Y.: admetsar 2.0: web-service for prediction and optimization of chemical
  admet properties. Bioinformatics  (2018)

\bibitem{yap2011padel}
Yap, C.W.: Padel-descriptor: An open source software to calculate molecular
  descriptors and fingerprints. Journal of computational chemistry
  \textbf{32}(7),  1466--1474 (2011)

\bibitem{zhu2008combinatorial}
Zhu, H., Tropsha, A., Fourches, D., Varnek, A., Papa, E., Gramatica, P., Oberg,
  T., Dao, P., Cherkasov, A., Tetko, I.V.: Combinatorial qsar modeling of
  chemical toxicants tested against tetrahymena pyriformis. Journal of chemical
  information and modeling  \textbf{48}(4),  766--784 (2008)

\end{thebibliography}

\bibliographystyle{splncs04}

\end{document}